\documentclass[conference]{IEEEtran}
\IEEEoverridecommandlockouts

\newif\ifarxiv
\arxivtrue

\usepackage{cite}
\usepackage{amsmath,amssymb}
\usepackage[T1]{fontenc}
\usepackage[utf8]{inputenc}
\usepackage{graphicx}
\usepackage{booktabs}
\usepackage{array}
\usepackage{algorithm}
\usepackage{xcolor}

\usepackage{url}
\usepackage{microtype}
\graphicspath{{figures/}}

\newcommand{\sig}[1]{\textsuperscript{#1}}

\begin{document}
\bstctlcite{BSTcontrol}

\title{Route, Don't Fix: Regime-Dependent Decoding Correction and a
Trajectory-Gated Router for Reliable Clinical LLM Answer Selection}

\author{%
\IEEEauthorblockN{Zeyu Dong\textsuperscript{*}} 
\IEEEauthorblockA{Crestwood Preparatory College\\
Toronto, Canada\\
dongzeyu123@outlook.com}
\and
\IEEEauthorblockN{Benjamin Wang\textsuperscript{*}}
\IEEEauthorblockA{St.\ Theresa of Lisieux Catholic High School\\
Toronto, Canada\\
benrw78@gmail.com}
\and
\IEEEauthorblockN{Joyee W. Jin\textsuperscript{\textdagger}}
\IEEEauthorblockA{Department of Computer Science\\
University of Toronto\\
Toronto, Canada\\
joyee.jin@mail.utoronto.ca}
\ifarxiv
\thanks{\copyright~2026 IEEE. Personal use of this material is permitted.
Permission from IEEE must be obtained for all other uses, in any current or
future media, including reprinting/republishing this material for advertising
or promotional purposes, creating new collective works, for resale or
redistribution to servers or lists, or reuse of any copyrighted component of
this work in other works. Accepted at the 2026 Third International Conference
on Artificial Intelligence for Medicine, Health and Care (AIxMHC).}
\fi
}

\IEEEaftertitletext{%
\vspace*{-2em}
\noindent\parbox{\textwidth}{\centering\footnotesize
\textsuperscript{*}These authors contributed equally and are co-first authors.\\
\textsuperscript{\textdagger}Supervising author.
}\par
\vspace{1em}
}

\maketitle

\begin{abstract}
Large language models (LLMs) are often deemed unsafe for clinical question answering
because of their tendency to hallucinate. Current methods for reducing
hallucination, such as retrieval augmentation, fine-tuning, and external verifier
models, are difficult to deploy in a clinical setting because each requires new
infrastructure that clinical governance must separately approve. Additionally,
each may add unwanted latency or extra model calls to every query. Inference-time methods remove the need for that
infrastructure by acting while the next token is being selected, using the model's
own internal logit signals. Four such methods (DoLa, DeLTa, chain-of-verification,
and inference-time intervention) reduce hallucination by applying one
\emph{fixed} transformation to every question. We show that a fixed transformation is the wrong granularity. A
corrector that improves LLM accuracy by about ten points on a truthfulness stress
test achieves only negligible gains over the greedy baseline when the same
corrector is applied to clinical multiple-choice benchmarks. On those benchmarks,
instruction tuning has already concentrated nearly all of the output probability
on one answer option by the final layer, eliminating the need for a
corrector. This concentration leaves a low output entropy and provides a motive for
applying a router between correction and greedy decoding. Our method, ALTAS, reads the terminal entropy and a late-layer linearity statistic
($R^2$) from one forward pass and chooses per question between greedy decoding and
a late-layer trajectory correction. In our method, no classifier, probe, or head
is trained; we instead perform operations on the logit values of each candidate
answer considered by the LLM. Our lightweight router adds only 6.5\% latency
overhead.
Applied to
every question, the correction improves TruthfulQA over greedy at 3B and 8B by 11.4
and 10.0 points respectively ($p<10^{-10}$). Gated per question, ALTAS keeps 8.3 to 9.5 of those
points while holding MedQA, PubMedQA, and MedHallu inside a do-no-harm band of one
point (none significant). Our method passes verification sweeps over the frozen
thresholds, the scoring rule, and the domain label.
\end{abstract}


\section{Introduction}
Large language models are increasingly proposed for clinical question answering,
triage, and literature interpretation. In those settings, a confidently stated but
unsupported answer can be dangerous or potentially
fatal~\cite{thirunavukarasu2023llm,vasey2022decideai}. Clinical mitigations
exist, but each is expensive, since retrieval-augmented
generation~\cite{lewis2020rag}, sampling-based
self-verification~\cite{manakul2023selfcheckgpt}, and supervised truthfulness
probes add a corpus, labels, latency, or extra model calls that governance
cannot always approve.

A lighter family of methods avoids that overhead by operating at decoding time on
a single forward pass; DoLa~\cite{chuang2024dola}, DeLTa~\cite{delta2025},
chain-of-verification (CoVe)~\cite{dhuliawala2023cove}, and inference-time
intervention (ITI)~\cite{li2023iti} are four such methods, although all four
assume that a fixed correction suits \emph{every} question.

Our locked evaluation shows that assumption failing (Table~\ref{tab:regime}). A
corrector worth about ten points on TruthfulQA cannot be distinguished from greedy
decoding on MedQA and PubMedQA, because on those benchmarks the instruction-tuned
base model has already concentrated its output distribution on one answer. In
other words, correction helps in one regime, while outside that regime the safest
action is to fall back to greedy decoding.

Falling back to greedy decoding is safe only if the model can tell which regime a question is in,
and telling the regimes apart takes one forward pass, in which the model reads its
terminal entropy and the linearity of its late-layer logit trajectory. With those
two readings, the decoder applies the corrector where the model's own trajectory
predicts that the corrector will help and falls back to greedy decoding
everywhere else. We
contribute ALTAS
(Sec.~\ref{sec:method}), a regime-aware decoder whose entropy and $R^2$ tests
choose per question between greedy decoding and a late-layer trajectory
correction, with no probe, classifier, or head trained (Table~\ref{tab:diff}).
We also contribute an $R^2$ viability test (Sec.~\ref{sec:r2}) that
ensures a model's late-layer logit trajectories are linear enough for the
late-layer extrapolation to be viable. Our third
contribution is
a leakage-free evaluation protocol over four benchmarks, with paired
significance tests, a per-question oracle, and a sensitivity analysis
(Sec.~\ref{sec:sens}).

\section{Related Work}
\label{sec:cands}
ALTAS routes among established decoding interventions, treating them as a pool
of candidates instead of competitors. Within that pool, \emph{layer-contrast
decoding} exploits how token probabilities evolve with depth, so that DoLa
contrasts a mature layer against a premature one~\cite{chuang2024dola},
generalizing contrastive decoding~\cite{li2023contrastive}, while DeLTa
extrapolates the same trajectory toward the converged
distribution~\cite{delta2025}. Both are token-level maps with no notion of the
question.

\emph{Sampling-based verification} estimates factuality from agreement across
stochastic samples (SelfCheckGPT~\cite{manakul2023selfcheckgpt}) or asks the model
to check its own draft (CoVe~\cite{dhuliawala2023cove}). Checking the model's own
generated answer yields a consistency-based estimate of its factuality, although
it requires $k$ extra generations per question and weakens in domains where the
model cannot verify itself, such as specialized clinical content. Clinical content is also where \emph{activation steering} such
as ITI~\cite{li2023iti} needs help, because ITI moves hidden states along a
truthful direction learned from labeled probe data, which requires supervision
and a per-model fit and which applies unconditionally once installed.

The fundamental difference is conditioning (Table~\ref{tab:diff}). Every method
above is a fixed map $f$ applied to all questions; ALTAS computes
$g(x)\in\{\text{greedy},f\}$ per question from the model's own late-layer
logit-trajectory geometry, using no samples, no labels, and no external judge.
Conditioning also separates ALTAS from existing routers, which work at a coarser
grain; RouteLLM~\cite{ong2024routellm}, for example, learns from preference data
\emph{which model} should answer, whereas ALTAS routes among \emph{decoding
strategies inside one fixed model} on one forward pass. To our knowledge, such
per-question, intra-model routing is unexplored.

\begin{table}[t]
\centering
\caption{Comparison of ALTAS with other methods.}
\label{tab:diff}
\scriptsize
\setlength{\tabcolsep}{3pt}
\begin{tabular}{l l l l c}
\toprule
Method & Reads & Extra work & Per-model fit & Per-$q$ \\
\midrule
DoLa~\cite{chuang2024dola} & layer contrast & none & none & no \\
DeLTa~\cite{delta2025} & layer trajectory & none & none & no \\
SelfCheckGPT~\cite{manakul2023selfcheckgpt} & sample agreement & $k$ samples & none & no \\
CoVe~\cite{dhuliawala2023cove} & self-check draft & extra passes & none & no \\
ITI~\cite{li2023iti} & steered states & none & labeled probe & no \\
\textbf{ALTAS} & $H^{(N)}$, late $R^2$, tag & 1 prefill (6.5\%) & none & \textbf{yes} \\
\bottomrule
\end{tabular}
\end{table}

\section{Method}
\label{sec:method}
Six design choices fix the procedure, and each is detailed below. The gate reads
three quantities: the terminal entropy $H^{(N)}$, the model-level late-layer
linearity $\bar R^2$, and a clinical/general tag (Sec.~\ref{sec:feat}). Here $\bar R^2$ is measured from half depth (layers $\lfloor N/2 \rfloor \!:\! N$) on 50 candidates, while the
correction anchors at layer 7 and fits layers $14\!:\!N$ on 200 candidates
(Sec.~\ref{sec:alta}). The decision is three scalar threshold comparisons with
nothing trained (Algorithm~\ref{alg:altas}), and the correction itself is
Eqs.~(\ref{eq:corr})--(\ref{eq:delta}) with $\alpha_c=\alpha_e=0.3$. The three
constants are $\tau_{R^2}=0.50$, $\tau_H=0.5$ nats, and a cosine acceptance of
0.65, the first two calibrated without labels on 15 prompts
(Sec.~\ref{sec:router}). Routing adds one extra prefill, or 6.5\% of greedy
latency (Sec.~\ref{sec:cost}).

\subsection{Trajectory correction}
\label{sec:alta}
\begin{figure}[t]
\centering
\includegraphics[width=\linewidth]{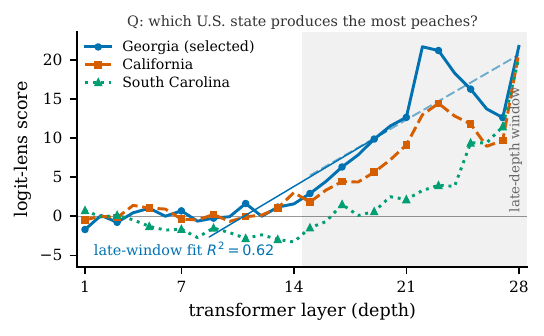}
\caption{Logit lens at every layer of Llama-3.2-3B on the TruthfulQA item
``Which U.S.\ state produces the most peaches?'' Candidates are entangled early
and separate only inside the shaded window (layers 15--28), where \emph{Georgia}
keeps gaining support while \emph{California} and \emph{South Carolina} plateau.
The late-window linearity (here 0.62) is the feature the gate reads.}
\label{fig:traj}
\end{figure}

Let the model have $N$ layers, and let the shared vocabulary head project each
layer's hidden state into logits, giving a per-layer next-token distribution. At
decoding step $t$, the corrected terminal logit is
\begin{equation}
\label{eq:corr}
z'_t = z^{(N)}_t + g_t\,\Delta_t,\qquad
g_t=\mathrm{clip}\!\big(H^{(N)}_t/3,\,0,\,1\big),
\end{equation}
\begin{equation}
\label{eq:delta}
\Delta_t = \alpha_c\big(z^{(N)}_t - z^{(7)}_t\big)
         + \alpha_e\big(\hat z_t - z^{(N)}_t\big),
\end{equation}
where $H^{(N)}_t$ is the terminal entropy in nats, the first term is a DoLa-style
contrast against a fixed anchor at layer 7, and $\hat z_t$ extrapolates a per-token
least-squares fit over layers $14\!:\!N$ by one virtual layer, restricted to the
200 highest-logit candidates. The anchor and the window were set once on
Llama-3.2-3B ($N=28$), where the anchor sits at a quarter of depth, below where
candidates separate, and the window opens at half depth, where the trajectory is
close to linear (Fig.~\ref{fig:traj}). Both correction weights, $\alpha_c$ and
$\alpha_e$, are set to 0.3, following DoLa~\cite{chuang2024dola} and
DeLTa~\cite{delta2025}, while safety comes from the entropy gate $g_t$, which
falls to zero as the terminal distribution concentrates; because a zero gate
reduces the operator to greedy scoring, the correction cannot overwrite a
confidently resolved clinical answer. Note that the anchor and window are
\emph{absolute} layer indices and are therefore not depth-normalized, a choice
Sec.~\ref{sec:disc} returns to.

\subsection{Gate readings}
\label{sec:feat}
A single forward pass over the question prompt, keeping each layer's intermediate
hidden states, yields every quantity the gate needs. From half depth onward, the gate takes the 50 highest-logit candidates at the final position, fits each candidate's logit against depth, and
averages those fits into the late-layer linearity $R^2$, while the same pass also
gives the terminal entropy $H^{(N)}$. The calibration therefore uses only the
terminal entropy, the model-level $R^2$, and the clinical/non-clinical label.

\subsection{Decision rule}
\label{sec:router}
We calibrated the two constants used in Algorithm~\ref{alg:altas}: the viability
cutoff $\tau_{R^2}=0.50$, which the model's late-layer linearity must clear
before the correction is used at all, and the entropy threshold $\tau_H=0.5$
nats, which a question's terminal entropy must exceed before the correction is
applied. They were measured over 15 TruthfulQA prompts before evaluation, without
using any labels. Because that sample is small, Sec.~\ref{sec:sens} measures how
much the conclusions depend on these constants. A third constant, the cosine acceptance threshold of 0.65, governs scoring
only (Sec.~\ref{sec:cos}). Neither calibration probe reads a gold answer, since
both summarize internal logit-trajectory geometry, and no gold-answer label is
used in any routing decision, thus preventing data leakage.

\textbf{ALTAS is an inference-time method that requires no external training.}
It adds no parameters, learned probe, or
classifier, because the gate compares three scalars against three constants, and
those constants, together with the layer indices and weights of
Sec.~\ref{sec:alta}, form the complete specification, exposed in one configuration
file of the released implementation. The decision is binary, and the correction
fires on 73.8\%, 70.4\%, and 43.8\% of TruthfulQA items at 3B, 8B, and 14B, and on
no clinical item at any scale.

\begin{algorithm}[t]
\caption{ALTAS decides one question with three scalar comparisons, where the
correction reuses layer logits that each decoding step already computes.}
\label{alg:altas}
\hrule\vspace{3pt}
\footnotesize\noindent
\textbf{Input:} question $x$; domain tag $d$; model-level $\bar R^2$.

\noindent\textbf{1.}~Run one forward pass over the prompt; read $H^{(N)}$ and the late-layer
descriptors (Sec.~\ref{sec:feat}).\\
\textbf{2.}~\textbf{if} $\bar R^2 \ge \tau_{R^2}$ \textbf{and} $d=\text{general}$
\textbf{and} $H^{(N)}>\tau_H$ \textbf{then}\\
\hspace*{1.4em}decode with the correction of Sec.~\ref{sec:alta};\\
\textbf{3.}~\textbf{else} decode greedily, with output identical to the base model's.
\vspace{3pt}\hrule
\end{algorithm}

The clinical/non-clinical label is load-bearing, and every run reported here takes
that label from the benchmark, so MedQA, PubMedQA, and MedHallu items carry
\texttt{clinical} and TruthfulQA items carry \texttt{general}. When the label is
absent, the implementation falls back to a lexical keyword test whose effect
Sec.~\ref{sec:sens} measures.

\subsection{One $R^2$ cutoff per model family}
\label{sec:r2}
Whether late-layer extrapolation is viable depends on how linear the trajectory
is, which the model-level $R^2$ summarizes. Measured without labels over 15
TruthfulQA prompts, that model-level statistic is numerically stable across scale
(Fig.~\ref{fig:r2}a), so one cutoff serves both Llama models without per-scale
retuning. Qwen2.5-14B marks the limit of the test, however, because there the same
reading of 0.53 clears the cutoff and labels the correction viable even though the
correction is not viable (Table~\ref{tab:main}); late-layer linearity therefore
captures correction viability \emph{within a family}, despite the absence of that
link \emph{across architectures}.

\subsection{Cost}
\label{sec:cost}
Rather than loading a second model, calling a verifier, or adding a decoding
step, ALTAS adds only one prefill: the two routing signals are read in a single
extra forward pass over the prompt before generation begins. With that prefill,
Llama-3.2-3B in bfloat16 on one RTX 5060 laptop GPU, over 20 TruthfulQA prompts
(three warm-ups, median prompt 64 tokens, 80 new tokens), gives a feature-pass
median of 377\,ms against a greedy median of 5.82\,s, so routing adds 6.5\% to
the cheap greedy branch. Corrected decoding is the expensive branch, because the
operator projects the late layers through the vocabulary head at every step,
taking 9.78\,s, or 1.68$\times$ greedy, while peak memory is unchanged within
0.03\,GB. Since the corrector fires on no clinical question, a clinical workload
incurs only the feature pass.

\subsection{Cosine answer scoring}
\label{sec:cos}
All protocols are scored identically, in that each candidate answer is embedded
with a sentence encoder (\texttt{all-MiniLM-L6-v2}) and compared to the reference by cosine
similarity, at an acceptance threshold of 0.65. That threshold never decides a
single-token clinical answer (a letter or yes/no/maybe), because option and polarity
matching resolve those answers before similarity is consulted. Cosine agreement
there coincides with exact match to the gold answer (Sec.~\ref{sec:sens}). Free-form TruthfulQA and
MedHallu answers receive a graded score under the same rule, which makes every
comparison paired at the question level. Accuracy is reported over the full
benchmark, with unscorable generations counted as incorrect instead of dropped.

\section{Experiments}
\subsection{Setup}
We evaluate three instruction-tuned models (Llama-3.2-3B,
Llama-3.1-8B~\cite{meta2024llamahf}, and Qwen2.5-14B~\cite{qwen2024card}) on four
benchmarks, TruthfulQA~\cite{lin2022truthfulqa}, MedHallu~\cite{pandit2025medhallu},
MedQA~\cite{jin2021medqa}, and PubMedQA~\cite{jin2019pubmedqa} (test-set sizes
in Table~\ref{tab:main}). We compare four protocols, namely greedy
decoding, the DoLa$+$DeLTa ablation, the always-on correction, and ALTAS.
Correction and ALTAS are compared to greedy with a paired bootstrap of 5000
replicates and exact McNemar tests~\cite{mcnemar1947,efron1994bootstrap} at 95\%.
We mark $p<10^{-3}$ as \sig{***}, $p<10^{-2}$ as \sig{**}, $p<0.05$ as \sig{*},
and everything else as \sig{ns} (not significant). The low PubMedQA 14B greedy value is a base-model
quirk, since Qwen2.5-14B answers ``maybe'' on 244 of 500 items against a gold rate
of 10\%.

At 14B, the always-on correction degenerates on MedQA and PubMedQA for a mechanical
reason, namely that the operator reads intermediate layers through the tied output
head, a logit-lens transfer whose reliability varies across
architectures~\cite{belrose2023tunedlens}. That transfer holds on Llama but breaks
on Qwen2.5, collapsing single-token generation. It is our operator that fails at
14B rather than decoding-time correction in general, because
SLED~\cite{zhang2024sled} does improve factuality on Qwen.

\subsection{Main results}
\begin{figure*}[t]
\centering
\includegraphics[width=\textwidth]{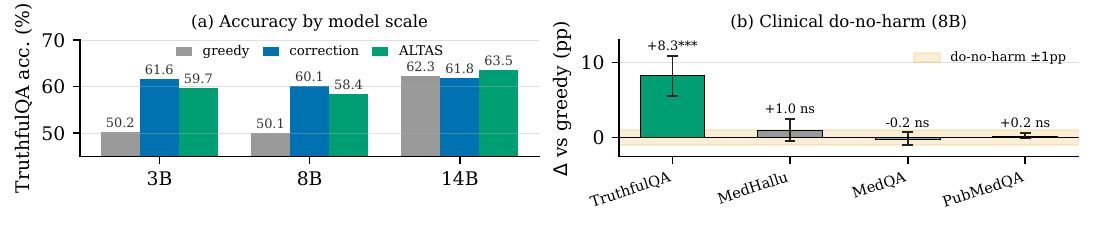}
\caption{(a) TruthfulQA accuracy at each model size. Both the correction and
ALTAS gain about ten points at 3B and 8B, but only 1.2 (\sig{ns}) at 14B. (b) How
much ALTAS changes accuracy relative to greedy decoding at 8B. TruthfulQA gains
8.3 points (\sig{***}), while every clinical benchmark stays within one point of
greedy. Bars are bootstrap 95\% confidence intervals, and stars are exact
McNemar tests.}
\label{fig:acc}
\end{figure*}

Table~\ref{tab:main} and Fig.~\ref{fig:acc} report the full grid, where we see a
one-directional pattern. On TruthfulQA the always-on correction improves accuracy by
11.4 and 10.0 points at the two Llama scales, and the gated decoder by 9.5 and
8.3, all four gains highly significant. Those gains collapse at 14B, where the always-on correctors
fall to or below greedy and ALTAS reaches a non-significant 1.2. Across the three
clinical benchmarks, meanwhile, every ALTAS delta is non-significant and inside a
one-point do-no-harm band at every scale.

At 14B, corrected free-form output begins coherent and degrades in the tail, so
whole-answer cosine scoring tolerates the degradation while single-token scoring
does not. Free-form MedHallu accordingly still shows always-on gains of 5.9 and 4.4
(\sig{***}/\sig{**}), which we read as forgiving scoring of a partly degraded
output instead of a valid correction, consistent with TruthfulQA-14B showing no
gain ($-$0.5, \sig{ns}). Gating removes those gains (ALTAS $-$0.8, \sig{ns}).
On clinical single-token benchmarks the same correction collapses outright, taking
MedQA and PubMedQA to 0.9\% and 36.6\%. This failure is sidestepped by the router
as it defaults to greedy.

\subsection{Regime dependence}
Table~\ref{tab:regime} isolates the argument at 8B. Applied everywhere, the
correction improves accuracy by 10.0 on TruthfulQA and 11.8 on MedHallu (both \sig{***}) while
staying inert on MedQA and PubMedQA at 0.6 and 1.0 (\sig{ns}). We observe here how
one fixed map can be strong in one regime and worthless in another. Gated per question, the same
operator transfers the TruthfulQA gain (8.3, \sig{***}) and stays inside the
do-no-harm band on clinical multiple-choice, although the price is MedHallu, where
the gate routes every item to greedy and keeps 1.0 of the 11.8 (\sig{ns}).

\subsection{Oracle gap}
\begin{figure*}[t]
\centering
\includegraphics[width=\textwidth]{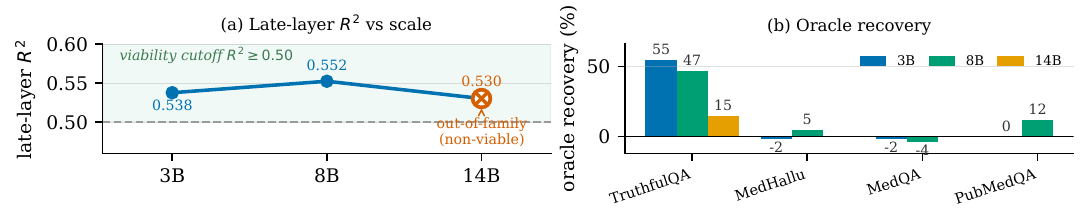}
\caption{(a) How linear each model's late-layer trajectories are. A single cutoff
of 0.50, measured on Llama 3B, correctly admits both Llama models without
retuning. Qwen2.5-14B also clears the cutoff, yet the correction fails there,
which marks the limit of the test. (b) How much of the available gain ALTAS
actually captures, measured against a perfect router that always picks the better
of the two actions. On TruthfulQA it captures 55\%, 47\%, and 15\% at 3B, 8B, and
14B, and almost none on the clinical benchmarks.}
\label{fig:r2}
\end{figure*}

A per-question oracle, which selects for each question whichever of the two
actions yields the correct answer, bounds what any gate could have achieved
(Fig.~\ref{fig:r2}b). Even on TruthfulQA, where ALTAS works best, it captures only
55\% of that available gain at 3B and 47\% at 8B, leaving most of it unclaimed. On
clinical multiple-choice it captures almost none. At 14B, the oracle reaches 70.3
on TruthfulQA while ALTAS captures 15\%. This gap shows that the bottleneck is
deciding which action to take, not the corrector itself. Sec.~\ref{sec:sens} next asks whether those
results depend on the constants we froze.

\begin{table*}[t]
\centering
\caption{Accuracy (\%, $\uparrow$) under cosine scoring, with change versus greedy
and McNemar stars as in Sec.~IV-A. Correction applies our operator to every
question; ALTAS gates it per question.}
\label{tab:main}
\footnotesize
\begin{tabular}{l c c c c c c}
\toprule
Benchmark & $n$ & Scale & Greedy & DoLa$+$DeLTa & Correction & ALTAS \\
\midrule
TruthfulQA & 817 & 3B & 50.2 & 61.2\,(+11.0)\sig{***} & \textbf{61.6\,(+11.4)\sig{***}} & \textbf{59.7\,(+9.5)\sig{***}} \\
 & 817 & 8B & 50.1 & 61.4\,(+11.4)\sig{***} & \textbf{60.1\,(+10.0)\sig{***}} & \textbf{58.4\,(+8.3)\sig{***}} \\
 & 817 & 14B & 62.3 & 60.0\,($-$2.3)\sig{ns} & 61.8\,($-$0.5)\sig{ns} & 63.5\,(+1.2)\sig{ns} \\
\midrule
MedHallu & 1000 & 3B & 46.6 & 56.6\,(+10.0)\sig{***} & \textbf{57.3\,(+10.7)\sig{***}} & 46.3\,($-$0.3)\sig{ns} \\
 & 1000 & 8B & 47.0 & 58.7\,(+11.7)\sig{***} & \textbf{58.8\,(+11.8)\sig{***}} & 48.0\,(+1.0)\sig{ns} \\
 & 1000 & 14B & 53.9 & 59.8\,(+5.9)\sig{***} & 58.3\,(+4.4)\sig{**} & 53.1\,($-$0.8)\sig{ns} \\
\midrule
MedQA & 1273 & 3B & 52.5 & 54.0\,(+1.6)\sig{*} & 53.4\,(+0.9)\sig{ns} & 52.4\,($-$0.1)\sig{ns} \\
 & 1273 & 8B & 57.9 & 58.4\,(+0.5)\sig{ns} & 58.5\,(+0.6)\sig{ns} & 57.7\,($-$0.2)\sig{ns} \\
 & 1273 & 14B & 66.8 & 1.4\,($-$65.4)\sig{***} & 0.9\,($-$65.8)\sig{***} & 66.8\,(+0.1)\sig{ns} \\
\midrule
PubMedQA & 500 & 3B & 73.8 & 72.6\,($-$1.2)\sig{ns} & 72.6\,($-$1.2)\sig{ns} & 73.8\,(+0.0)\sig{ns} \\
 & 500 & 8B & 77.8 & 77.0\,($-$0.8)\sig{ns} & 78.8\,(+1.0)\sig{ns} & 78.0\,(+0.2)\sig{ns} \\
 & 500 & 14B & 52.2 & 36.6\,($-$15.6)\sig{***} & 32.6\,($-$19.6)\sig{***} & 52.2\,(+0.0)\sig{ns} \\
\bottomrule
\end{tabular}
\end{table*}

\begin{table}[t]
\centering
\caption{Change versus greedy at 8B, with paired bootstrap 95\% CIs in brackets
and stars as in Table~\ref{tab:main}.}
\label{tab:regime}
\footnotesize
\setlength{\tabcolsep}{4pt}
\begin{tabular}{l c c}
\toprule
Benchmark & Correction (fixed) & ALTAS (routed) \\
\midrule
TruthfulQA & $+10.0$\,\sig{***}\,{\scriptsize[7.2, 12.9]} & $+8.3$\,\sig{***}\,{\scriptsize[5.6, 10.9]} \\
MedHallu   & $+11.8$\,\sig{***}\,{\scriptsize[9.1, 14.6]} & $+1.0$\,\sig{ns}\,{\scriptsize[$-$0.4, 2.5]} \\
MedQA      & $+0.6$\,\sig{ns}\,{\scriptsize[$-$0.5, 1.8]}  & $-0.2$\,\sig{ns}\,{\scriptsize[$-$1.0, 0.7]} \\
PubMedQA   & $+1.0$\,\sig{ns}\,{\scriptsize[0.0, 2.2]}    & $+0.2$\,\sig{ns}\,{\scriptsize[0.0, 0.6]} \\
\bottomrule
\end{tabular}
\end{table}


\section{Sensitivity Analysis}
\label{sec:sens}
Three constants were frozen on calibration sets of 15 prompts before evaluation.
This section asks whether the conclusions rest on those constants, on the
scoring rule, or on the domain label. Because every gate feature, both decoding
arms, and every generation sit in the locked logs over 3590 questions, each sweep
below is a counterfactual over recorded evidence and requires no new run.
Reconstructing ALTAS at the frozen settings reproduces the routed fraction
and the logged accuracy to within 1.3.

\emph{The entropy threshold sits on a plateau.}
Sweeping $\tau_H$ across [0, 3.0] drives the routed fraction from near-total
coverage to near zero and traces the TruthfulQA gain smoothly with it
(Table~\ref{tab:sens}), and any value in [0, 1.0] still leaves a gain above 5 at
both Llama scales. The frozen threshold therefore sits on a plateau instead of a peak.

\emph{The viability cutoff tolerates a wide range of values, but its 15-prompt
estimator does not.} The
cutoff enters as one scalar comparison against a
model-level $R^2$ of 0.5378, 0.5525, and 0.5303. Every routing decision here is
unchanged for any cutoff inside [0, 0.53]. Fragility lives in the estimator
instead, because when the probe set is resampled 20,000 times from the 817 logged
per-question values, a 15-prompt probe at 8B lands inside [0.557, 0.606] and never
falls below the cutoff. At 3B, the same probe lands inside [0.454, 0.517]
and falls below the cutoff in 82\% of draws. Calibration at 8B is therefore
stable, while at 3B it is one probe draw from non-viability. A deployment
should use far more than 15 prompts.

\emph{A deeper measurement window would have caught the 14B failure.} The
deployed statistic reads from half depth and clears the cutoff at
14B with 0.5303; read over the top 30\% of layers instead, the statistic returns
0.4436 over 50 prompts and 0.4393 pooled over all logged items, below the cutoff.
The measurement window therefore decides whether a viability test detects the
Qwen2.5 collapse.

\emph{Free-form gains survive cosine thresholds from 0.50 to 0.90.} Re-scoring
every locked generation at cosine thresholds from 0.50 to 0.90
reproduces every sign, each logged accuracy to within 1.1, and each logged delta to
within 0.4. The 8B always-on MedHallu gain stays positive across the range (from
1.3 to 12.5); the ALTAS TruthfulQA gain is positive throughout and peaks near 0.75 at 10.0, above the 8.2 this re-scoring returns at the frozen 0.65; and the ALTAS MedHallu delta stays inside
1.3 everywhere. Clinical multiple-choice never raises the question, because
option and polarity matching decide MedQA and PubMedQA items before similarity at
rates of 99.9\% and 100.0\% respectively.

\emph{The fallback domain classifier reaches precision 0.99 and recall 0.82, with
a bounded worst case.} Scored against benchmark-level ground
truth over every logged question, the fallback lexical test reaches precision 0.99
and recall 0.82 for the clinical class. Recall is uneven, however, at 95.4\%,
96.2\%, and 57.4\% on MedQA, PubMedQA, and MedHallu respectively, because MedHallu
is held down by terse biomedical titles with no keyword. Substituting the lexical test for the
benchmark tag is the worst case for do-no-harm. Of the 59 MedQA items
the test mislabels at 14B, 5 also clear the entropy gate, moving benchmark accuracy
by $-$0.24; for PubMedQA the figures are 3 items and $-$0.60. MedHallu is the
exception in the opposite direction, where 337 items reach the corrector at 8B and
\emph{raise} accuracy by 3.4. The opposite error, wrongly flagging a general
question as clinical, is rarer still, affecting 24 TruthfulQA items at a loss of
at most 0.12.

\begin{table}[t]
\centering
\caption{Routed share and TruthfulQA change versus greedy as $\tau_H$ sweeps past
its frozen value of 0.5 (locked logs).}
\label{tab:sens}
\footnotesize
\setlength{\tabcolsep}{4pt}
\begin{tabular}{l c c c c}
\toprule
$\tau_H$ & routed 3B & TQA $\Delta$ 3B & routed 8B & TQA $\Delta$ 8B \\
\midrule
0.0            & 99.9\% & $+11.3$ & 99.8\% & $+9.7$ \\
\textbf{0.5}   & \textbf{73.8\%} & $\mathbf{+9.1}$ & \textbf{70.4\%} & $\mathbf{+8.0}$ \\
1.0            & 51.0\% & $+7.4$  & 44.1\% & $+5.4$ \\
1.5            & 24.8\% & $+3.6$  & 19.0\% & $+2.0$ \\
\bottomrule
\end{tabular}
\end{table}

\section{Discussion}
\label{sec:disc}
Two mechanisms explain the regime dependence. The first is \emph{RLHF entropy
collapse}, whereby instruction tuning sharpens the terminal distribution on
in-distribution tasks~\cite{ouyang2022rlhf}. On clinical multiple-choice,
most of the probability mass already sits on one option. That concentration leaves
a correction nothing to act on, since the best it can do is match greedy
while risking overwriting a correct answer. This is precisely the condition
the entropy gate detects. TruthfulQA, by contrast, is built from misconceptions the
model is pulled toward, and it therefore carries higher terminal entropy that gives the
correction room to work. The second mechanism is \emph{domain restriction of
verification} as self-checking is limited in clinical content.

Neither mechanism explains the Qwen2.5-14B collapse, for which there are two
candidate causes. One is logit-lens decodability, that is, how faithfully a
layer's hidden state can be read as a token distribution through the final output
head, which varies by architecture~\cite{belrose2023tunedlens}. The other is our own
operator, whose anchor and window are absolute layer indices. Over 48
layers, the window opens at 29\% of depth against 50\% for Llama-3.2-3B.
Depth-normalizing those indices is therefore the next experiment in future work.

The oracle gap shows that the next advance is a better gate instead of a stronger
corrector, and MedHallu proves this as the always-on correction adds
11.8 there ($p<10^{-17}$) yet ALTAS routes all 1000 items to greedy at both 3B and
8B. That same abstention also makes deployment simple. ALTAS stands down whenever
a clinical answer is already resolved, leaving the output \emph{identical} to the
base model's on every question it does not flag.

\section{Limitations}
Since decoding is deterministic (greedy, seed 42), there is no sampling or seed
variance, and only a sub-point MedHallu residual from non-bit-identical free-form
passes remains.

More specifically, there are four boundaries. First, the transfer of
the $R^2$ cutoff is demonstrated on \emph{two} model sizes inside \emph{one} model family, and
the third scale is a different architecture on which transfer fails. Second,
clinical do-no-harm rests on a benchmark-supplied clinical tag; the lexical
fallback recalls clinical items at 82\%, and Sec.~\ref{sec:sens} bounds the worst
case for these three benchmarks alone (at $-$0.6). Third, the calibration probes
are small (15 prompts), and the consequence is asymmetric, harmless at 8B and one
draw from flipping the verdict at 3B. Fourth, cosine scoring is graded for
free-form answers, and the residual risk there concerns magnitude.

Additionally, against a per-question oracle, the gated decoder
recovers at most 55\% of the available gain, which places the open problem in
routing.

Finally, public benchmarks remain proxies for clinical workflows. ALTAS is a
reliability layer but not yet a decision-support system which requires external
validation.

\section{Conclusion}
Decoding-time hallucination correction is regime-dependent because the fixed
corrector that helps on a truthfulness stress test is inert on clinical
multiple-choice and destructive on one architecture. ALTAS responds by routing instead of fixing; its entropy-and-$R^2$ gate chooses per question
between greedy decoding and a late-layer trajectory correction under a
calibration that transfers within a model family. It thereby delivers the
TruthfulQA gain without a significant accuracy reduction on the evaluated
clinical benchmarks, and this result holds at every threshold value we tested. ALTAS, however, does not close the per-question oracle gap, showing that the next advance would belong to the gate instead of the corrector.

\bibliographystyle{IEEEtran}
\bibliography{refs}

\end{document}